\documentclass{article}
\usepackage{spconf}

\usepackage{cite}
\usepackage{amsmath,amssymb,amsfonts}
\usepackage{graphicx}
\usepackage{textcomp}
\usepackage[svgnames,x11names]{xcolor}

\usepackage{array}
\usepackage{mathtools}
\usepackage{tcolorbox}
\usepackage{color}
\usepackage{theorem}
\usepackage{amssymb}
\usepackage{cite,hyperref}
\usepackage{cases}
\usepackage{url}
\usepackage{algpseudocode}
\usepackage{algorithm}
\usepackage{enumitem}
\usepackage{multirow}
\usepackage{hhline}
\input{styles/mysymbol.sty}

\usepackage{tikz}
\usetikzlibrary{shapes,arrows}
\usepackage{moresize}
\usepackage{pgfplots}
\usepackage{wrapfig}
\pgfplotsset{compat=1.17}\usepgfplotslibrary{statistics} 
\pgfplotstableset{col sep=comma}

\newcommand \rmss[1] {{\rm \scriptscriptstyle #1}}

\pgfplotsset{compat=1.17}
\pgfplotstableset{col sep=comma}
\tikzset{every mark/.append style={scale=1.5, solid}, font=\footnotesize}
\pgfplotsset{
    width=1.05\textwidth,
    legend style={
        font=\scriptsize ,  
        inner xsep=1pt,
        inner ysep=1pt,
        nodes={inner sep=1pt}},
    legend cell align=left,
	every axis/.append style={line width=0.5pt},
	every axis plot/.append style={line width=.9pt},
    every axis y label/.append style={yshift=-3pt}
}

\begin{document}
\ninept

\title{Learning Time-Varying Turn-Taking Behavior in Group Conversations}

\name{Madeline Navarro, Lisa O'Bryan, and Santiago Segarra
\thanks{Funding for this project was provided by the Army Research Institute for the Behavioral and Social Sciences (Grant W911NF-22-1-0226). The views, opinions, and/or findings contained in this article are those of the authors and shall not be construed as an official Department of the Army position, policy, or decision, unless so designated by other
documents.
        Emails:  
        \{\href{mailto:nav@rice.edu}{nav}, 
        \href{mailto:obryan@rice.edu}{obryan},
        \href{mailto:segarra@rice.edu}{segarra}\}@rice.edu }
}
\address{Rice University, Houston, TX, USA}

\maketitle

\begin{abstract}
    We propose a flexible probabilistic model for \emph{predicting turn-taking patterns in group conversations} based solely on \emph{individual characteristics} and \emph{past speaking behavior}.
    Many models of conversation dynamics cannot yield insights that generalize beyond a single group.
    Moreover, past works often aim to characterize speaking behavior through a universal formulation that may not be suitable for all groups.
    We thus develop a generalization of prior conversation models that predicts speaking turns among individuals in any group based on their individual characteristics, that is, personality traits, and prior speaking behavior.
    Importantly, our approach provides the novel ability to \emph{learn} how speaking inclination varies based on when individuals last spoke.
    We apply our model to synthetic and real-world conversation data to verify the proposed approach and characterize real group interactions.
    Our results demonstrate that previous behavioral models may not always be realistic, motivating our data-driven yet theoretically grounded approach.
\end{abstract}


\section{Introduction}
\label{S:intro}


Understanding how individuals behave in group settings is valuable for both analyzing human behavior and improving communication-based tools.
The ability to model group interactions can facilitate understanding of how they emerge from group composition variables, support the design of harmonious and effective teams based on team member compatibility, and enable the prediction of future group dynamics, informing potential interventions~\cite{bell2018TeamcompositionABCs,obryan2020HowApproachesAnimal,kozlowski2018unpackingTeamProcess}. 
Moreover, insights into complex human interactions can contribute to building language models that are more realistic, understandable, or helpful~\cite{zhou2024IsThisReal}.

While major developments in natural language processing have contributed to simulating realistic, text-based communication~\cite{kim2023SODA}, there is still much to explore regarding how turn-taking behavior evolves in conversations and how team composition shapes these dynamics.
Many past works that model conversation dynamics have aimed to capture universal conversation mechanics, which are generalizable but may not be suitable for all groups~\cite{stasser1991Speakingturnsfacetoface,obryan2025MLSPEAK}. 
Other existing approaches have limited investigation to the behavior of a single group and linked differences in speaking behaviors to specific individuals, limiting generalizability to other  groups~\cite{parker1988Speakingturnssmall,basu2001LearningHumanInteractions,padilha2002SimulationSmallGroup}. 
When it comes to the study of individual differences in speaking behaviors, traditional approaches generally involve interpretable yet simple tools such as those based on linear regression.
For example, studies have examined how personality traits, such as dominance, correlate with individual speaking time~\cite{littlepage1995input} or how mean team-level trait values, such as extraversion, correlate with the total number of utterances made by the team~\cite{macht2014StructuralModelsExtraversion}.
However, due to the interactive nature of conversational data~\cite{sacks1974simplest}, methods are required that take into account its dynamic and interdependent characteristics~\cite{kuljanin2024AdvancingOrganizationalScience,bleidorn2019UsingMachineLearning}. 
Unfortunately, while data-driven tools demonstrate the ability to link individual characteristics to more complex behavioral patterns, the majority of such works implement black-box models, from which it is challenging to draw theoretical conclusions~\cite{yarkoni2017ChoosingPredictionExplanation,stachl2020PersonalityResearchAssessment}. 

We therefore propose a flexible parametric model for conversational turn-taking that predicts speaking patterns for groups of individuals.
In particular, we generalize a well-founded statistical model that, at any point in a conversation, returns the probability that any individual will be the next speaker based on their personality traits and prior turn-taking behavior~\cite{stasser1991Speakingturnsfacetoface}. 
Our model is built by fitting behavioral parameters to observed conversation data among groups with known traits~\cite{kimble1988DominanceArguingMixedsex,cuperman2009bigfivepredictors}. 
We not only return scores reflecting the personalized speaking tendencies of any individual, but, differently from prior work~\cite{stasser1991Speakingturnsfacetoface,obryan2022ConversationalTurntakingStochastic,obryan2025MLSPEAK}, we also provide the novel ability to \emph{learn} how speaking inclination changes based on how recently individuals last spoke.
Thus, our model fitting process adapts to dynamic speaking patterns that may differ across group settings.
This allows us to empirically evaluate if existing assumptions on how behavior varies over time align with real-world data~\cite{stasser1991Speakingturnsfacetoface,parker1988Speakingturnssmall}. 
Moreover, beyond predictive ability, our use of machine learning constrained to theoretically grounded behavioral models yields interpretable characterizations of the relationship between personality and speaking behavior~\cite{obryan2025MLSPEAK}. 
Our contributions are as follows.
\begin{itemize}[left= 8pt .. 15pt, noitemsep]
    \item[1.] We propose a flexible probabilistic model of turn-taking behavior for individuals in groups dependent upon their traits and conversation history, with the novel ability to learn how speaking inclination varies with the gap since individuals last spoke.
    \item[2.] We verify our model by learning from both synthetic and real-world conversational data, evaluated by measuring our ability to predict conversations both globally and with an emphasis on rarer, more complex turn-taking patterns.
    \item[3.] We compare our learned time-varying speaking behavior to existing assumptions, allowing us to investigate how real conversations evolve and thus assess the realism of common assumptions.
\end{itemize}

\section{Problem statement}
\label{S:problem}

We seek a model that predicts turn-taking patterns in a group conversation given observed traits of the group members.
To this end, we consider a set of $N$ individuals with measured traits collected in the vector $\bbx \in \reals^{N}$, where $x_i \in \reals$ denotes the trait of the $i$-th individual.
These may correspond to any measured value associated with an individual, such as level of expertise on a group task; in this work, we focus on $\bbx$ representing personality traits.
We exemplify our work with scalar-valued traits for simplicity, but extension to the multivariate setting is feasible and straightforward~\cite{obryan2025MLSPEAK}.
We further observe a sequence $\ccalC := \{ s(t) \}_{t=1}^T$ of $T$ turns, where $s(t) \in [N] := \{1,2,\dots,N\}$ represents which individual speaks at the $t$-th turn.
We aim to obtain the likelihood that each individual will speak at the $t$-th turn given traits $\bbx$ and conversation history $\ccalH_{t-1} := \{ s(j) \}_{j=1}^{t-1} \subset \ccalC$ up to turn $t-1$.
To this end, we consider the following parametric approach to characterize speaking likelihoods based on a well-established conversational model~\cite{stasser1991Speakingturnsfacetoface}.
At each turn $t$, the probability of member $i$ speaking is $\mbP[s(t) = i] = u_i(t) / (\bbone^\top \bbu(t))$, where $\ccalU := \{ \bbu(t) \}_{t=1}^T$ such that $\bbu(t) \in \reals_+^N$ contains the time-varying speaking likelihoods of all $N$ individuals.
For a given member $i \in [N]$, we define
\alna{
    u_i(t)
    &~:=~&
    \Big[ 
        \pi_i + d_i w\big(\delta_i(t) \big) 
    \Big] 
    \mbI\{ \delta_i(t) > 1 \},
\label{eq:score}}
where $\mbI\{\cdot\}$ is the indicator function and $\bbpi,\bbd \in \reals_+^N$ denote scores of speaking tendencies.
The \emph{inherent} tendency to speak for member $i$ is $\pi_i$, while $d_i$ reflects speaker \emph{memory}, as $d_i$ scales the time-dependent term in~\eqref{eq:score}.
The \emph{proclivity} of a speaker $w: \mbZ \rightarrow \reals_{+}$ represents how speaking inclination varies with the number of turns that have elapsed since an individual last spoke (previously termed the `memory function'~\cite{obryan2025MLSPEAK}), where $w(\delta) = 0$ for any $\delta \leq 0$, and $\delta_i(t) \in \naturals$ denotes the number of turns between $t$ and the last turn in $\ccalH_{t-1}$ spoken by $i$, with $\delta_i(t) = t - \max\{ j : s(j) = i, j < t \}$ if member $i$ has spoken at least once, otherwise, $\delta_i(t) = -\infty$. 
The first turn therefore depends only on the inherent scores $\bbu(1) = \bbpi$.
Thus, larger values of $w$ in~\eqref{eq:score} increase the chance that member $i$ will speak, which may vary with how recently $i$ last spoke.
Observe that by the condition $\delta_i(t) > 1$, individuals cannot speak for consecutive turns, as we define a turn ending when another member speaks.

To discover how personality influences conversational behavior, we implement the model in~\eqref{eq:score} to predict speaking patterns from personality traits.
Past works often assume $w(t) = e^{-b\delta_i(t)}$ for some $b > 0$, reflecting the greater tendency of recent speakers to speak again, with this likelihood decaying exponentially as the number of turns since they last spoke increases~\cite{obryan2022ConversationalTurntakingStochastic, stasser1991Speakingturnsfacetoface, obryan2025MLSPEAK}.
Differently, our representation in~\eqref{eq:score} allows for alternative patterns of time-varying behavior which can be learned directly from turn-taking data.
Indeed, while $w(t) = e^{-b\delta_i(t)}$ provides a natural theoretical model, it may not reflect different patterns of speaking in all scenarios.
Beyond modeling capabilities, existing work exploiting frameworks similar to~\eqref{eq:score} are typically employed to analyze the behavior of a given group based on observed conversation data, which cannot be generalized to unseen groups~\cite{stasser1991Speakingturnsfacetoface,obryan2022ConversationalTurntakingStochastic,parker1988Speakingturnssmall}.
While progress has been made towards generalizability~\cite{obryan2025MLSPEAK}, adapting to time-varying patterns that may differ across group settings has not yet been explored. 

\begin{figure*}[t]
    \centering
    \hspace{-.4cm}
    \begin{minipage}[b][][b]{.3\textwidth}
        \begin{tikzpicture}[baseline,scale=.9,trim axis left, trim axis right]

\pgfplotstableread{data/synth_exp_vanilla_boxplots_2025-09-08.csv}\expv
\pgfplotstableread{data/synth_exp_turntype_boxplots_2025-09-08.csv}\expt
\pgfplotstableread{data/synth_sig_vanilla_boxplots_2025-09-08.csv}\sigv
\pgfplotstableread{data/synth_sig_turntype_boxplots_2025-09-08.csv}\sigt


\definecolor{madblack}{HTML}{5c5f77}
\definecolor{madred}{HTML}{d20f39}
\definecolor{madgreen}{HTML}{40a02b}
\definecolor{madyellow}{HTML}{df8e1d}
\definecolor{madblue}{HTML}{1e66f5}
\definecolor{madmagenta}{HTML}{ea76cb}
\definecolor{madcyan}{HTML}{ea76cb}
\definecolor{madgray}{HTML}{acb0be}

\begin{axis}[
    ymajorgrids,
    boxplot/draw direction=y,
    xtick={2,8},
    xticklabels={Loss $\ell$,Loss $\ell_{\rmss{turn}}$},
    xlabel={(a) Proclivity $w_{\exp}$},
    xmin=-.75,
    xmax=10.75,
    ymax=1.42,
    legend style={
        at={(.5, 1.02)},
        anchor=south},
    legend columns=5,
    width=210,
    height=140,
    label style={font=\small},
    tick label style={font=\small}
    ]

    \addplot+[
        boxplot,
        boxplot/every median/.style={madblack,solid, line width=2pt},
        boxplot/every whisker/.style={madblack,solid, line width=1.6pt},
        boxplot/every box/.style={madblack,solid, line width=1.6pt},
        every mark/.append style={mark=o,madblack},
        boxplot/draw position=0.0,
        forget plot
    ] table[ col sep=comma, y=true_exp ] {\expv};
     \addplot+[
        boxplot,
        boxplot/every median/.style={madred,solid, line width=2pt},
        boxplot/every whisker/.style={madred,solid, line width=1.6pt},
        boxplot/every box/.style={madred,solid, line width=1.6pt},
        every mark/.append style={mark=o,madred},
        boxplot/draw position=1.0,
        forget plot
    ] table[ col sep=comma, y=ours_exp ] {\expv};
     \addplot+[
        boxplot,
        boxplot/every median/.style={madgreen,solid, line width=2pt},
        boxplot/every whisker/.style={madgreen,solid, line width=1.6pt},
        boxplot/every box/.style={madgreen,solid, line width=1.6pt},
        every mark/.append style={mark=o,madgreen},
        boxplot/draw position=2.0,
        forget plot
    ] table[ col sep=comma, y=exp_exp ] {\expv};
     \addplot+[
        boxplot,
        boxplot/every median/.style={madblue,solid, line width=2pt},
        boxplot/every whisker/.style={madblue,solid, line width=1.6pt},
        boxplot/every box/.style={madblue,solid, line width=1.6pt},
        every mark/.append style={mark=o,madblue},
        boxplot/draw position=3.0,
        forget plot
    ] table[ col sep=comma, y=nomem_exp ] {\expv};
     \addplot+[
        boxplot,
        boxplot/every median/.style={madyellow,solid, line width=2pt},
        boxplot/every whisker/.style={madyellow,solid, line width=1.6pt},
        boxplot/every box/.style={madyellow,solid, line width=1.6pt},
        every mark/.append style={mark=o,madyellow},
        boxplot/draw position=4.0,
        forget plot
    ] table[ col sep=comma, y=mem_exp ] {\expv};

    \addplot+[
        boxplot,
        boxplot/every median/.style={madblack,solid,line width=2pt},
        boxplot/every whisker/.style={madblack,solid,line width=1.6pt},
        boxplot/every box/.style={madblack,solid,fill=madblack!40,line width=1.6pt},
        every mark/.append style={mark=*,madblack,fill=madblack!40},
        boxplot/draw position=6,
        forget plot
    ] table[ col sep=comma, y=true_exp ] {\expt};
     \addplot+[
        boxplot,
        boxplot/every median/.style={madred,solid,line width=2pt},
        boxplot/every whisker/.style={madred,solid,line width=1.6pt},
        boxplot/every box/.style={madred,solid,fill=madred!40,line width=1.6pt},
        every mark/.append style={mark=*,madred,fill=madred!40},
        boxplot/draw position=7,
        forget plot
    ] table[ col sep=comma, y=ours_exp ] {\expt};
     \addplot+[
        boxplot,
        boxplot/every median/.style={madgreen,solid,line width=2pt},
        boxplot/every whisker/.style={madgreen,solid,line width=1.6pt},
        boxplot/every box/.style={madgreen,solid,fill=madgreen!40,line width=1.6pt},
        every mark/.append style={mark=*,madgreen,fill=madgreen!40},
        boxplot/draw position=8,
        forget plot
    ] table[ col sep=comma, y=exp_exp ] {\expt};
     \addplot+[
        boxplot,
        boxplot/every median/.style={madblue,solid,line width=2pt},
        boxplot/every whisker/.style={madblue,solid,line width=1.6pt},
        boxplot/every box/.style={madblue,solid,fill=madblue!40,line width=1.6pt},
        every mark/.append style={mark=*,madblue,fill=madblue!40},
        boxplot/draw position=9,
        forget plot
    ] table[ col sep=comma, y=nomem_exp ] {\expt};
     \addplot+[
        boxplot,
        boxplot/every median/.style={madyellow,solid,line width=2pt},
        boxplot/every whisker/.style={madyellow,solid,line width=2pt},
        boxplot/every box/.style={madyellow,solid,fill=madyellow!40,line width=2pt},
        every mark/.append style={mark=*,madyellow,fill=madyellow!40},
        boxplot/draw position=10,
        forget plot
    ] table[ col sep=comma, y=mem_exp ] {\expt};

    \addlegendimage{only marks, mark=square*, mark options={fill=madblack,draw=madblack}}
    \addlegendentry{True~~~~}
    \addlegendimage{only marks, mark=square*, mark options={fill=madred,draw=madred}}
    \addlegendentry{PRO~~~~}
    \addlegendimage{only marks, mark=square*, mark options={fill=madgreen,draw=madgreen}}
    \addlegendentry{EXP~~~~}
    \addlegendimage{only marks, mark=square*, mark options={fill=madblue,draw=madblue}}
    \addlegendentry{NM~~~~}
    \addlegendimage{only marks, mark=square*, mark options={fill=madyellow,draw=madyellow}}
    \addlegendentry{HM}
    
    
\end{axis}

\end{tikzpicture}
    \end{minipage}
    \hspace{.5cm}
    \begin{minipage}[b][][b]{.3\textwidth}
        \begin{tikzpicture}[baseline,scale=.9,trim axis left, trim axis right]

\pgfplotstableread{data/synth_exp_vanilla_boxplots_2025-09-08.csv}\expv
\pgfplotstableread{data/synth_exp_turntype_boxplots_2025-09-08.csv}\expt
\pgfplotstableread{data/synth_sig_vanilla_boxplots_2025-09-08.csv}\sigv
\pgfplotstableread{data/synth_sig_turntype_boxplots_2025-09-08.csv}\sigt


\definecolor{madblack}{HTML}{5c5f77}
\definecolor{madred}{HTML}{d20f39}
\definecolor{madgreen}{HTML}{40a02b}
\definecolor{madyellow}{HTML}{df8e1d}
\definecolor{madblue}{HTML}{1e66f5}
\definecolor{madmagenta}{HTML}{ea76cb}
\definecolor{madcyan}{HTML}{ea76cb}
\definecolor{madgray}{HTML}{acb0be}

\begin{axis}[
    ymajorgrids,
    boxplot/draw direction=y,
    xtick={2,8},
    xticklabels={Loss $\ell$,Loss $\ell_{\rmss{turn}}$},
    xlabel={(b) Proclivity $w_{\rm sig}$},
    xmin=-.75,
    xmax=10.75,
    ymax=1.42,
    legend style={
        at={(.5, 1.02)},
        anchor=south},
    legend columns=5,
    width=210,
    height=140,
    label style={font=\small},
    tick label style={font=\small}
    ]

    \addplot+[
        boxplot,
        boxplot/every median/.style={madblack,solid, line width=2pt},
        boxplot/every whisker/.style={madblack,solid, line width=1.6pt},
        boxplot/every box/.style={madblack,solid, line width=1.6pt},
        every mark/.append style={mark=o,madblack},
        boxplot/draw position=0.0,
        forget plot
    ] table[ col sep=comma, y=true_sig ] {\sigv};
     \addplot+[
        boxplot,
        boxplot/every median/.style={madred,solid, line width=2pt},
        boxplot/every whisker/.style={madred,solid, line width=1.6pt},
        boxplot/every box/.style={madred,solid, line width=1.6pt},
        every mark/.append style={mark=o,madred},
        boxplot/draw position=1.0,
        forget plot
    ] table[ col sep=comma, y=ours_sig ] {\sigv};
     \addplot+[
        boxplot,
        boxplot/every median/.style={madgreen,solid, line width=2pt},
        boxplot/every whisker/.style={madgreen,solid, line width=1.6pt},
        boxplot/every box/.style={madgreen,solid, line width=1.6pt},
        every mark/.append style={mark=o,madgreen},
        boxplot/draw position=2.0,
        forget plot
    ] table[ col sep=comma, y=exp_sig ] {\sigv};
     \addplot+[
        boxplot,
        boxplot/every median/.style={madblue,solid, line width=2pt},
        boxplot/every whisker/.style={madblue,solid, line width=1.6pt},
        boxplot/every box/.style={madblue,solid, line width=1.6pt},
        every mark/.append style={mark=o,madblue},
        boxplot/draw position=3.0,
        forget plot
    ] table[ col sep=comma, y=nomem_sig ] {\sigv};
     \addplot+[
        boxplot,
        boxplot/every median/.style={madyellow,solid, line width=2pt},
        boxplot/every whisker/.style={madyellow,solid, line width=1.6pt},
        boxplot/every box/.style={madyellow,solid, line width=1.6pt},
        every mark/.append style={mark=o,madyellow},
        boxplot/draw position=4.0,
        forget plot
    ] table[ col sep=comma, y=mem_sig ] {\sigv};

    \addplot+[
        boxplot,
        boxplot/every median/.style={madblack,solid,line width=2pt},
        boxplot/every whisker/.style={madblack,solid,line width=1.6pt},
        boxplot/every box/.style={madblack,solid,fill=madblack!40,line width=1.6pt},
        every mark/.append style={mark=*,madblack,fill=madblack!40},
        boxplot/draw position=6,
        forget plot
    ] table[ col sep=comma, y=true_sig ] {\sigt};
     \addplot+[
        boxplot,
        boxplot/every median/.style={madred,solid,line width=2pt},
        boxplot/every whisker/.style={madred,solid,line width=1.6pt},
        boxplot/every box/.style={madred,solid,fill=madred!40,line width=1.6pt},
        every mark/.append style={mark=*,madred,fill=madred!40},
        boxplot/draw position=7,
        forget plot
    ] table[ col sep=comma, y=ours_sig ] {\sigt};
     \addplot+[
        boxplot,
        boxplot/every median/.style={madgreen,solid,line width=2pt},
        boxplot/every whisker/.style={madgreen,solid,line width=1.6pt},
        boxplot/every box/.style={madgreen,solid,fill=madgreen!40,line width=1.6pt},
        every mark/.append style={mark=*,madgreen,fill=madgreen!40},
        boxplot/draw position=8,
        forget plot
    ] table[ col sep=comma, y=exp_sig ] {\sigt};
     \addplot+[
        boxplot,
        boxplot/every median/.style={madblue,solid,line width=2pt},
        boxplot/every whisker/.style={madblue,solid,line width=1.6pt},
        boxplot/every box/.style={madblue,solid,fill=madblue!40,line width=1.6pt},
        every mark/.append style={mark=*,madblue,fill=madblue!40},
        boxplot/draw position=9,
        forget plot
    ] table[ col sep=comma, y=nomem_sig ] {\sigt};
     \addplot+[
        boxplot,
        boxplot/every median/.style={madyellow,solid,line width=2pt},
        boxplot/every whisker/.style={madyellow,solid,line width=1.6pt},
        boxplot/every box/.style={madyellow,solid,fill=madyellow!40,line width=1.6pt},
        every mark/.append style={mark=*,madyellow,fill=madyellow!40},
        boxplot/draw position=10,
        forget plot
    ] table[ col sep=comma, y=mem_sig ] {\sigt};

    \addlegendimage{only marks, mark=square*, mark options={fill=madblack,draw=madblack}}
    \addlegendentry{True~~~~}
    \addlegendimage{only marks, mark=square*, mark options={fill=madred,draw=madred}}
    \addlegendentry{PRO~~~~}
    \addlegendimage{only marks, mark=square*, mark options={fill=madgreen,draw=madgreen}}
    \addlegendentry{EXP~~~~}
    \addlegendimage{only marks, mark=square*, mark options={fill=madblue,draw=madblue}}
    \addlegendentry{NM~~~~}
    \addlegendimage{only marks, mark=square*, mark options={fill=madyellow,draw=madyellow}}
    \addlegendentry{HM}
    
    
\end{axis}

\end{tikzpicture}
    \end{minipage}
    \hspace{.5cm}
    \begin{minipage}[b][][b]{.3\textwidth}
        \begin{tikzpicture}[baseline,scale=.9,trim axis left, trim axis right]

\pgfplotstableread{data/real_extra_vanilla_boxplots_2025-09-11.csv}\realv
\pgfplotstableread{data/real_extra_turntype_boxplots_2025-09-11.csv}\realt


\definecolor{madblack}{HTML}{5c5f77}
\definecolor{madred}{HTML}{d20f39}
\definecolor{madgreen}{HTML}{40a02b}
\definecolor{madyellow}{HTML}{df8e1d}
\definecolor{madblue}{HTML}{1e66f5}
\definecolor{madmagenta}{HTML}{ea76cb}
\definecolor{madcyan}{HTML}{ea76cb}
\definecolor{madgray}{HTML}{acb0be}

\begin{axis}[
    ymajorgrids,
    boxplot/draw direction=y,
    xtick={2.5,7.5},
    xticklabels={Loss $\ell$,Loss $\ell_{\rmss{turn}}$},
    xlabel={(c) Real data},
    xmin=.25,
    xmax=9.75,
    legend style={
        at={(.5, 1.02)},
        anchor=south},
    legend columns=5,
    width=210,
    height=140,
    label style={font=\small},
    tick label style={font=\small}
    ]

     \addplot+[
        boxplot,
        boxplot/every median/.style={madred,solid, line width=2pt},
        boxplot/every whisker/.style={madred,solid, line width=1.6pt},
        boxplot/every box/.style={madred,solid, line width=1.6pt},
        every mark/.append style={mark=o,madred},
        boxplot/draw position=1.0,
        forget plot
    ] table[ col sep=comma, y=ours_real ] {\realv};
     \addplot+[
        boxplot,
        boxplot/every median/.style={madgreen,solid, line width=2pt},
        boxplot/every whisker/.style={madgreen,solid, line width=1.6pt},
        boxplot/every box/.style={madgreen,solid, line width=1.6pt},
        every mark/.append style={mark=o,madgreen},
        boxplot/draw position=2.0,
        forget plot
    ] table[ col sep=comma, y=exp_real ] {\realv};
     \addplot+[
        boxplot,
        boxplot/every median/.style={madblue,solid, line width=2pt},
        boxplot/every whisker/.style={madblue,solid, line width=1.6pt},
        boxplot/every box/.style={madblue,solid, line width=1.6pt},
        every mark/.append style={mark=o,madblue},
        boxplot/draw position=3.0,
        forget plot
    ] table[ col sep=comma, y=nomem_real ] {\realv};
     \addplot+[
        boxplot,
        boxplot/every median/.style={madyellow,solid, line width=2pt},
        boxplot/every whisker/.style={madyellow,solid, line width=1.6pt},
        boxplot/every box/.style={madyellow,solid, line width=1.6pt},
        every mark/.append style={mark=o,madyellow},
        boxplot/draw position=4.0,
        forget plot
    ] table[ col sep=comma, y=mem_real ] {\realv};

     \addplot+[
        boxplot,
        boxplot/every median/.style={madred,solid,line width=2pt},
        boxplot/every whisker/.style={madred,solid,line width=1.6pt},
        boxplot/every box/.style={madred,solid,fill=madred!40,line width=1.6pt},
        every mark/.append style={mark=*,madred,fill=madred!40},
        boxplot/draw position=6,
        forget plot
    ] table[ col sep=comma, y=ours_real ] {\realt};
     \addplot+[
        boxplot,
        boxplot/every median/.style={madgreen,solid,line width=2pt},
        boxplot/every whisker/.style={madgreen,solid,line width=1.6pt},
        boxplot/every box/.style={madgreen,solid,fill=madgreen!40,line width=1.6pt},
        every mark/.append style={mark=*,madgreen,fill=madgreen!40},
        boxplot/draw position=7,
        forget plot
    ] table[ col sep=comma, y=exp_real ] {\realt};
     \addplot+[
        boxplot,
        boxplot/every median/.style={madblue,solid,line width=2pt},
        boxplot/every whisker/.style={madblue,solid,line width=1.6pt},
        boxplot/every box/.style={madblue,solid,fill=madblue!40,line width=1.6pt},
        every mark/.append style={mark=*,madblue,fill=madblue!40},
        boxplot/draw position=8,
        forget plot
    ] table[ col sep=comma, y=nomem_real ] {\realt};
     \addplot+[
        boxplot,
        boxplot/every median/.style={madyellow,solid,line width=2pt},
        boxplot/every whisker/.style={madyellow,solid,line width=1.6pt},
        boxplot/every box/.style={madyellow,solid,fill=madyellow!40,line width=1.6pt},
        every mark/.append style={mark=*,madyellow,fill=madyellow!40},
        boxplot/draw position=9,
        forget plot
    ] table[ col sep=comma, y=mem_real ] {\realt};

    \addlegendimage{only marks, mark=square*, mark options={fill=madred,draw=madred}}
    \addlegendentry{PRO~~~~}
    \addlegendimage{only marks, mark=square*, mark options={fill=madgreen,draw=madgreen}}
    \addlegendentry{EXP~~~~}
    \addlegendimage{only marks, mark=square*, mark options={fill=madblue,draw=madblue}}
    \addlegendentry{NM~~~~}
    \addlegendimage{only marks, mark=square*, mark options={fill=madyellow,draw=madyellow}}
    \addlegendentry{HM}
    
    
\end{axis}

\end{tikzpicture}
    \end{minipage}
    \hspace{-1.4cm}
    \vspace{-.2cm}
    \caption{\small{
    (a)~Testing loss for each baseline when $w = w_{\rmss{exp}}$. 
    The left five boxplots are measured via $\ell$ in~\eqref{eq:loss}, the right five via $\ell_{\rmss{turn}}$ in~\eqref{eq:ttloss}.
    (b)~Testing loss for each baseline when $w = w_{\rmss{sig}}$. 
    (c)~Testing loss for each baseline using real-world conversation data. 
    }}
    \vspace{-.4cm}
\label{f:boxplots}
\end{figure*}
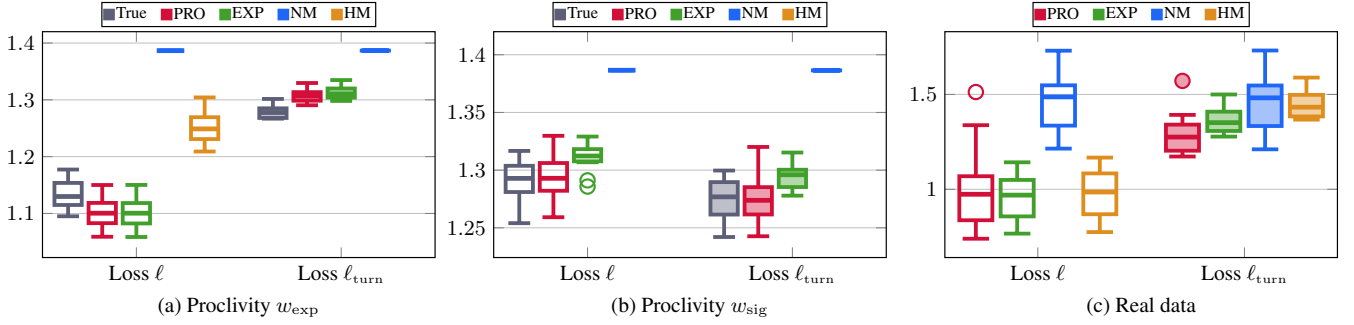

\section{Methodology}
\label{S:method}

Our goal is to predict the likelihoods $\bbu(t)$ in~\eqref{eq:score} for any turn $t \in \naturals$ and any group of arbitrary size $N$ given their traits $\bbx\in\reals^N$ and conversation history $\ccalH_{t-1} = \{ s(j) \}_{j = 1}^{t-1}$.
To this end, we train models $f,g:\reals \rightarrow \reals_+$ and $\nu:\mbZ \rightarrow \reals_+$ such that, for any $x\in\reals$, $\delta\in\mbZ$, 
\alna{
    a( x, \delta; \bbtheta )
    &~:=~&
    \Big[
        f( x; \bbtheta_1 ) + g( x; \bbtheta_2) \nu( \delta; \bbtheta_3 )
    \Big]
    \mbI\{ \delta > 1 \}
\label{eq:model}}
for learnable parameters $\bbtheta = (\bbtheta_1, \bbtheta_2, \bbtheta_3)$.
We may then let $\hat{u}_i(t) = a(x_i,\delta_i(t);\bbtheta)$ to obtain speaking likelihood estimates $\hat{\ccalU} = \{ \hbu(t) \}_{t=1}^T$,
where $\hat{\pi}_i = f(x_i;\bbtheta_1)$ and $\hat{d}_i = g(x_i;\bbtheta_2)$ predict inherent and memory scores, respectively, and $\nu$ approximates speaking proclivity with $\nu(\delta;\bbtheta_3) = 0$ if $\delta \leq 0$.
To obtain $f$, $g$, and $\nu$, we consider $G$ training groups, with traits $\bbx^{(g)} \in \reals^{N_g}$ belonging to the $g$-th group, among whom we observe a conversation $\ccalC^{(g)} = \{ s^{(g)}(t) \}_{t=1}^{T_g}$.
We learn the parameters $\bbtheta$ by maximizing the likelihood of each $\ccalC^{(g)}$ given $\bbx^{(g)}$ via
\alna{
    \min_{\bbtheta}
    ~
    \sum_{g=1}^G
    \ell( \hat{\ccalU}^{(g)}, \ccalC^{(g)} )
    ~~{\rm s.t.}~~
    &&
    \hat{u}_i^{(g)}(t) = a\big( x_i^{(g)}, \delta^{(g)}_i(t); \bbtheta \big)
&\nonumber\\[-.2cm]&
    &&
    \forall ~ i\in[N_g], ~ t\in[T_g], ~ g\in[G],
\label{eq:prob}}
where, for any $\ccalU = \{ \bbu(t) \}_{t=1}^T$ and $\ccalC \in [N]^T$, the loss function
\alna{
    \ell( \ccalU, \ccalC )
    ~:=~
    -\sum_{t=1}^T
    \log \frac{ u_{s(t)}(t) }{ \bbone^\top \bbu(t) }
\label{eq:loss}}
returns the negative log-likelihood of conversation $\ccalC$ given $\ccalU$ with $\bbu(t) \in \reals_+^N$.
While our approach supports several architectures, in this paper we let each of the components $f$, $g$, and $\nu$ be multilayer perceptrons (MLPs) with a sigmoidal nonlinearity ${\rm sig}(x) := ( 1 + e^{-x} )^{-1}$ at the output since we aim to predict nonnegative values $\bbu(t)$.
We solve~\eqref{eq:prob} via block coordinate descent~\cite{bertsekas1999nonlinear}, where we alternately update $(\bbtheta_1,\bbtheta_2)$ and $\bbtheta_3$ via gradient descent, precluding the need to compute gradients of the bilinear term in~\eqref{eq:model}, which can lead to instabilities.
Further details on both our architecture and simulations can be found in our GitHub repository\footnote{Link to repository: \href{https://github.com/mn51/mlspeak_proclivity}{https://github.com/mn51/mlspeak-proclivity}}.

By learning parameters of an interpretable and theoretically founded conversation model~\eqref{eq:model}, we enjoy benefits distinct from other machine learning techniques, elaborated below.
While sequential models such as recurrent neural networks (RNNs) are historical staples for time-series predictions~\cite{sutskever2014SequenceSequenceLearning}, these do not allow generalizability across groups of different sizes.
Indeed, tools that fix input dimensionality cannot describe turn-taking behavior across arbitrary groups.
For example, Markov models have shown success in reflecting conversational patterns, albeit only for a single observed group~\cite{parker1988Speakingturnssmall}.
In contrast, transformers allow arbitrary input and output dimensions and have been applied to create realistic sequences of data~\cite{vaswani2017AttentionAllYou}.
However, such architectures are also highly uninterpretable and therefore unlikely to soon be widely adopted for behavioral analyses~\cite{stachl2020PersonalityResearchAssessment}.
In contrast, we follow similar paths as interdisciplinary approaches in science-informed machine learning, which train models restricted to well-understood constraints, such as theoretical rules or physical restrictions~\cite{ghaseminejad2020Physicsinspired}. 

Solving~\eqref{eq:prob} allows us to perform predictions on any set of individuals with known traits.
First, given the trait $x_i^{*}$ of any new individual, we may predict scores $\hat{\pi}_i^{*}$ and $\hat{d}_i^{*}$ via $f$ and $g$, respectively, allowing us to characterize how behavior varies across personality traits.
Second, for a set of individuals with traits $\bbx^{*} \in \reals^{N^*}$ and conversation history $\ccalH^{*}_{t-1}$, we predict the most likely next speaker as $\hat{s}^{*}(t) = \argmax_i a(x_i^{*},\delta_i^{*}(t);\bbtheta)$ from~\eqref{eq:model}.
Third, we may generate realistic conversations from $\bbx^{*}$ by sampling speakers at each turn $s(t) \sim \hbu(t) / (\bbone^\top \hbu(t))$ with $\hat{u}_i(t) = a(x_i,\delta_i(t);\bbtheta)$.
Thus, our approach advances capabilities of turn-taking analysis, sharing the goals of our past work~\cite{obryan2025MLSPEAK}.
However, ML-SPEAK in~\cite{obryan2025MLSPEAK} did not learn a generic proclivity $\nu$ and instead assumed $w(\delta) = e^{-\delta/2}$, which we show may not be suited to all real-world conversations.

\subsection{Evaluating predictive ability}
\label{Ss:loss}

To assess the ability of our model to accurately describe speaking tendencies, we measure the \textit{test loss}, the loss for a conversation $\ccalC^*$ among a new testing group given traits $\bbx^*$.
To this end, we predict inherent and memory scores $\hbpi^*$, $\hbd^*$ from $\bbx^*$ and estimate speaking likelihoods $\hat{\ccalU}^*$ as in~\eqref{eq:score} with the learned proclivity $\hat{w} = \nu$.
Then, we compute the loss $\ell( \hat{\ccalU}^*, \ccalC^* )$ in~\eqref{eq:loss}, where lower values indicate that our model is well aligned with the dynamics in $\ccalC^*$.

In addition to $\ell$, we also propose an alternative loss, where each turn is weighted based on its \emph{class}, determined by the pattern of turns taken prior $\ccalH_t$.
Following ~\cite{stasser1991Speakingturnsfacetoface}, we define the following $K:=4$ classes for a given turn $t$: 
(i)~\textbf{Floor}: $s(t) = s(t-2)$;
(ii)~\textbf{Broken floor}: $s(t) \neq s(t-2)$, $s(t-1) = s(t-3)$;
(iii)~\textbf{Regain}: $s(t) = s(t-3) \neq s(t-1)$, $s(t-2) = s(t-4)$;
(iv)~\textbf{Nonfloor}: $\ccalH_t$ does not satisfy (i), (ii), or (iii).
Then, if $T_k$ is the number of turns in class $k \in [K]$ 
and $\gamma(\ccalH_t) := T / (K T_k )$ for turn $t$ in class $k$,
we define
\alna{
    \ell_{\rm \scriptscriptstyle turn}(\ccalU, \ccalC)
    ~:=~
    -\sum_{t=1}^T
    \gamma(\ccalH_t)
    \log \frac{ u_{s(t)}(t) }{ \bbone^\top \bbu(t) }.
\label{eq:ttloss}}
Therefore, turns in classes that occur more frequently (larger $T_k$) are given smaller weights.
As individuals who have recently spoken typically have a higher likelihood of speaking again, real-world conversations often contain a preponderance of floor turns~\cite{parker1988Speakingturnssmall,stasser1991Speakingturnsfacetoface}.
However, we wish to evaluate the ability of our model to predict a variety of patterns beyond those involving only two speakers.
Thus, while the loss $\ell$ captures the overall likelihood of a conversation, $\ell_{\rmss{turn}}$ measures how well a model can predict more complex and realistic patterns by weighting the likelihoods of each turn class equitably.

\section{Results}
\label{S:results}

We next apply our model~\eqref{eq:model} to synthetic and real-world data to predict speaking patterns.
In particular, we compare the following approaches to estimate proclivity $w$ and the scores $\bbpi, \bbd \in \reals_+^N$ of any group with traits $\bbx\in\reals^N$ and conversation $\ccalC = \{ s(t) \}_{t=1}^T$.
\begin{itemize}[left= 0pt .. 8pt, noitemsep]
    \item 
        \textbf{Proclivity (PRO)}: Our model~\eqref{eq:model} that predicts scores $\hat{\pi}_i^{\rmss{PRO}} = f(x_i;\bbtheta_1)$, $\hat{d}_i^{\rmss{PRO}} = g(x_i;\bbtheta_2)$ and learns proclivity $\hat{w}^{\rmss{PRO}}(\delta) = \nu(\delta;\bbtheta_3)$, with $\bbtheta = (\bbtheta_1, \bbtheta_2, \bbtheta_3)$ learned via~\eqref{eq:prob}; 
    \item
        \textbf{Exponential (EXP)}~\cite{obryan2025MLSPEAK}: A similar model that predicts scores $\hat{\pi}_i^{\rmss{EXP}} = f(x_i;\bbtheta_1)$, $\hat{d}_i^{\rmss{EXP}} = g(x_i;\bbtheta_2)$ for a \emph{fixed} proclivity $\hat{w}^{\rmss{EXP}}(\delta) = e^{-\delta/2}$, with $\bbtheta_1,\bbtheta_2$ learned via~\eqref{eq:prob}; 
    \item
        \textbf{No Memory (NM)}: A memoryless model that predicts scores $\hat{\pi}_i^{\rmss{NM}} = 1$, $\hat{d}_i^{\rmss{NM}} = 0$ and ignores proclivity $\hat{w}^{\rmss{NM}}(\delta) = 0$, 
    \item
        \textbf{High memory (HM)}: A high-memory model that predicts scores $\hat{\pi}_i^{\rmss{HM}} = 10^{-2}$, $\hat{d}_i^{\rmss{HM}} = 1$ with a fixed proclivity $\hat{w}^{\rmss{HM}}(\delta) = e^{-\delta/2}$.
\end{itemize}
Unlike our method \textbf{PRO}, \textbf{EXP} aligns with existing turn-taking analyses discussed in Section~\ref{S:problem}, which do not permit flexible modeling of how speaking inclination depends on the recency in which individuals last spoke~\cite{obryan2025MLSPEAK}.
The \textbf{NM} baseline predicts the next speaker uniformly at random with no consideration of conversation history.
Conversely, \textbf{HM} tends to select more recent speakers as the next predicted speaker, effectively ignoring inherent speaking tendencies.

\subsection{Synthetically generated conversations}

We first consider synthetic conversations to control the relationship between personality traits $\bbx$ and speaking likelihoods via $\bbpi,\bbd$.
We generate $15$ groups to be used for training and validation, each consisting of $5$ members, where we sample traits $x_i^{(g)} \sim [0.1,1]$ uniformly at random for each group $g \in [15]$.
We then obtain the inherent and memory scores $\bbpi^{(g)},\bbd^{(g)} \in \reals_+^N$ through the mappings
\alna{
    \pi_i^{(g)} = \sqrt{x_i^{(g)}},
    \quad
    d_i^{(g)} = 
    \frac{15e}{2}
    \left(
        \frac{ e^{-2(1.1 - x_i^{(g)}) } - e^{-2} }{ e^{-0.2} - e^{-2} }
        +
        \frac{1}{3}
    \right),
\label{eq:complexity}}
chosen to yield different nonlinear relationships such that $\pi_i^{(g)} \in [0,1]$ and $d_i^{(g)} \in [2.5e,10e]$.
Finally, we simulate a conversation per group based on $\bbpi^{(g)}$, $\bbd^{(g)}$ and a proclivity $w$ of choice, yielding sequences $\ccalC^{(g)} = \{ s^{(g)}(t) \}_{t=1}^{800}$.
We use $10$ groups to train the parameters of the \textbf{PRO} and \textbf{EXP} models via~\eqref{eq:prob}, and the remaining $5$ groups are set aside as validation data.

Furthermore, we generate $5$ testing groups in the same manner, each with 5 speakers and a conversation of $800$ turns, yielding traits $\bbx^{*(g)}$ and scores $\bbpi^{*(g)}, \bbd^{*(g)}$ for the $g$-th group, along with the conversation $\ccalC^{*(g)}$ simulated with the same proclivity $w$.
We evaluate each method by summing test losses across groups $\sum_{g} \ell(\hat{\ccalU}^{*(g)},\ccalC^{*(g)})$ via~\eqref{eq:loss} and $\sum_{g} \ell_{\rmss{turn}}(\hat{\ccalU}^{*(g)},\ccalC^{*(g)})$ via~\eqref{eq:ttloss}, where $\hat{\ccalU}^{*(g)}$ are the estimated likelihoods computed as in~\eqref{eq:score} from predictions $\hbpi^{*(g)}$, $\hbd^{*(g)}$, and $\hat{w}$.

We generate data for two types of proclivity,
\alna{
    w_{\rmss{exp}}(\delta)
    :=
    e^{-\delta / 2},
    \qquad
    w_{\rmss{sig}}(\delta)
    :=
    0.95 \,
    {\rm sig} \!
    \left( 10 - {\textstyle \frac{1}{2}} \delta \right),
\nonumber}
where $w_{\rmss{exp}}(\delta)$ aligns with assumptions in past works~\cite{stasser1991Speakingturnsfacetoface,obryan2022ConversationalTurntakingStochastic,obryan2025MLSPEAK}.
We further consider $w_{\rmss{sig}}(\delta)$, which also decreases as $\delta$ increases, yet as we will show, it renders the predictive task more challenging.

\vspace{.1cm}

\noindent\textbf{Exponentially decaying proclivity.}
First, we consider $10$ independent trials of data generated with $w = w_{\rmss{exp}}$.
The boxplots in Fig.~\ref{f:boxplots}a show the test losses summed over testing groups for $\ell$ in~\eqref{eq:loss} and $\ell_{\rmss{turn}}$ in~\eqref{eq:ttloss}.
Since we have ground truth scores $\bbpi^{*(g)}$, $\bbd^{*(g)}$ for each group and the true proclivity $w_{\rmss{exp}}$, we also compute test losses for the true scores $\ccalU^{*(g)}$, denoted \textbf{True} in Fig.~\ref{f:boxplots}a.
Note that Fig.~\ref{f:boxplots}a does not include \textbf{HM} for $\ell_{\rmss{turn}}$ as it attains significantly higher values (median approximately 1.56).
We find that \textbf{PRO} and \textbf{EXP} yield lower $\ell$ than \textbf{True}, but they do not outperform \textbf{True} for the class-based loss $\ell_{\rmss{turn}}$.
This indicates that \textbf{PRO} and \textbf{EXP} tend to predict frequently occurring floor turns more accurately but may struggle with more complex patterns, as expected for data created using a proclivity $w_{\rmss{exp}}$ that encourages floor turns.
Indeed, observe that \textbf{HM}, which assumes the correct proclivity $\hat{w}^{\rmss{HM}} = w_{\rmss{exp}}$, attains a lower $\ell$ than \textbf{NM}, but the more naive \textbf{NM} obtains a superior loss $\ell_{\rmss{turn}}$, implying that \textbf{HM} does not capture patterns accurately beyond floor turns.

\vspace{.1cm}

\noindent\textbf{Sigmoidal proclivity.}
We next consider the data sampled using a sigmoidal proclivity $w = w_{\rmss{sig}}$ for $10$ trials, with the test losses for each method shown in Fig.~\ref{f:boxplots}b.
Unlike for $w_{\rmss{exp}}$, we observe similar rankings between $\ell$ and $\ell_{\rmss{turn}}$.
First, note that \textbf{HM} with an inaccurate proclivity $\hat{w}^{\rmss{HM}}(\delta) = e^{-\delta/2}$ is not shown as it returns much higher losses $\ell$ and $\ell_{\rmss{turn}}$ than other methods (median values 1.69 and 1.55, respectively), where even the most naive approach \textbf{NM} outperforms \textbf{HM}.
However, \textbf{EXP} implements the same incorrect proclivity as \textbf{HM} $\hat{w}^{\rmss{EXP}}(\delta) = e^{-\delta/2}$, but \textbf{EXP} far outperforms \textbf{HM} as it additionally learns individual scores $\hbpi^{*(g)},\hbd^{*(g)}$.
Moreover, \textbf{PRO} exhibits superior loss to \textbf{EXP} for both $\ell$ and $\ell_{\rmss{turn}}$, as \textbf{PRO} attains values much closer to those of \textbf{True}.

\vspace{.1cm}
 
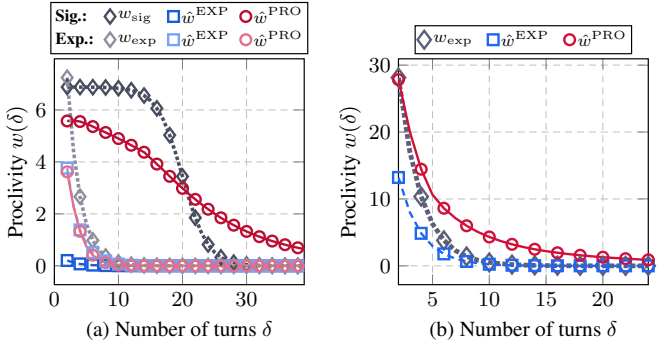
\begin{figure}[t]
    \centering
    \hspace{-.8cm}
    \begin{minipage}[b][][b]{.22\textwidth}
        \begin{tikzpicture}[baseline,scale=.9,trim axis left, trim axis right]

\pgfplotstableread{data/synth_exp_memfunc_2025-09-08.csv}\exptable
\pgfplotstableread{data/synth_sig_memfunc_2025-09-08.csv}\sigtable

\definecolor{madblack}{HTML}{5c5f77}
\definecolor{madred}{HTML}{d20f39}
\definecolor{madgreen}{HTML}{40a02b}
\definecolor{madyellow}{HTML}{df8e1d}
\definecolor{madblue}{HTML}{1e66f5}
\definecolor{madmagenta}{HTML}{ea76cb}
\definecolor{madcyan}{HTML}{ea76cb}
\definecolor{madgray}{HTML}{acb0be}

\begin{axis}[
    xlabel={(a) Number of turns $\delta$},
    xmin=0,
    xmax=39,
    ylabel={Proclivity $w(\delta)$},
    grid style=densely dashed,
    grid=both,
    legend style={
        at={(.5, 1.02)},
        anchor=south},
    legend columns=4,
    width=150,
    height=140,
    label style={font=\small},
    tick label style={font=\small}
    ]

    \addlegendimage{empty legend}
    \addplot[black!20!madblack, line width=1.5, densely dotted, forget plot] 
        table [x=turns, y=w_true] {\sigtable};
    \addplot[black!20!madblack, mark=diamond, line width=1, only marks, mark repeat=2, mark size=2] 
        table [x=turns, y=w_true] {\sigtable};

    \addplot[black!10!madblue, line width=1, densely dashed, forget plot] 
        table [x=turns, y=w_exp_avg] {\sigtable};
    \addplot[black!10!madblue, mark=square, line width=1, only marks, mark repeat=2, mark size=1.5] 
        table [x=turns, y=w_exp_avg] {\sigtable};

    \addplot[black!10!madred, line width=1, solid, forget plot] 
        table [x=turns, y=w_pred_avg] {\sigtable};
    \addplot[black!10!madred, mark=o, line width=1, only marks, mark repeat=2, mark size=1.5] 
        table [x=turns, y=w_pred_avg] {\sigtable};

    \addlegendimage{empty legend}
    \addplot[white!30!madblack, line width=1.5, densely dotted, forget plot] 
        table [x=turns, y=w_true] {\exptable};
    \addplot[white!30!madblack, mark=diamond, line width=1, only marks, mark repeat=2, mark size=2] 
        table [x=turns, y=w_true] {\exptable};

    \addplot[white!40!madblue, line width=1, densely dashed, forget plot] 
        table [x=turns, y=w_exp_avg] {\exptable};
    \addplot[white!40!madblue, mark=square, line width=1, only marks, mark repeat=2, mark size=1.5] 
        table [x=turns, y=w_exp_avg] {\exptable};

    \addplot[white!40!madred, line width=1, solid, forget plot] 
        table [x=turns, y=w_pred_avg] {\exptable};
    \addplot[white!40!madred, mark=o, line width=1, only marks, mark repeat=2, mark size=1.5] 
        table [x=turns, y=w_pred_avg] {\exptable};

    \legend{
        \textbf{Sig.:}~~~,
        $w_{\rmss{sig}}$~~~,
        $\hat{w}^{\rmss{EXP}}$~~~,
        $\hat{w}^{\rmss{PRO}}$,
        \textbf{Exp.:}~~~,
        $w_{\rmss{exp}}$~~~,
        $\hat{w}^{\rmss{EXP}}$~~~,
        $\hat{w}^{\rmss{PRO}}$,
    }
    
\end{axis}

\end{tikzpicture}
    \end{minipage}
    \hspace{.5cm}
    \begin{minipage}[b][][b]{.22\textwidth}
        \begin{tikzpicture}[baseline,scale=.9,trim axis left, trim axis right]

\pgfplotstableread{data/real_extra_memfunc_2025-09-11.csv}\realtable

\definecolor{madblack}{HTML}{5c5f77}
\definecolor{madred}{HTML}{d20f39}
\definecolor{madgreen}{HTML}{40a02b}
\definecolor{madyellow}{HTML}{df8e1d}
\definecolor{madblue}{HTML}{1e66f5}
\definecolor{madmagenta}{HTML}{ea76cb}
\definecolor{madcyan}{HTML}{ea76cb}
\definecolor{madgray}{HTML}{acb0be}

\begin{axis}[
    xlabel={(b) Number of turns $\delta$},
    xmin=2,
    xmax=24,
    ylabel={Proclivity $w(\delta)$},
    grid style=densely dashed,
    grid=both,
    legend style={
        at={(.5, 1.02)},
        anchor=south},
    legend columns=3,
    width=150,
    height=140,
    label style={font=\small},
    tick label style={font=\small}
    ]

    \addplot[madblack, line width=2, densely dotted, forget plot] 
        table [x=turns, y=w_true] {\realtable};
    \addplot[madblack, mark=diamond, line width=1, only marks, mark repeat=2, mark size=2.5] 
        table [x=turns, y=w_true] {\realtable};

    \addplot[madblue, line width=1, densely dashed, forget plot] 
        table [x=turns, y=w_exp_avg] {\realtable};
    \addplot[madblue, mark=square, line width=1, only marks, mark repeat=2, mark size=1.5] 
        table [x=turns, y=w_exp_avg] {\realtable};

    \addplot[madred, line width=1, solid, forget plot] 
        table [x=turns, y=w_pred_avg] {\realtable};
    \addplot[madred, mark=o, line width=1, only marks, mark repeat=2, mark size=1.5] 
        table [x=turns, y=w_pred_avg] {\realtable};
        
    \legend{
        $w_{\exp}$~~~~,
        $\hat{w}^{\rmss{EXP}}$~~~~,
        $\hat{w}^{\rmss{PRO}}$,
    }
    
\end{axis}

\end{tikzpicture}
    \end{minipage}
    \hspace{-1.3cm}
    \vspace{-.25cm}
    \caption{ \small{
    (a)~True versus learned proclivities for either a sigmoidal proclivity $w_{\rmss{sig}}$ or exponentially decaying proclivity $w_{\rmss{exp}}$.
    (b)~Learned proclivities for real-world conversation data.
    }}
    \vspace{-.5cm}
\label{f:memfuncs}
\end{figure}

\noindent\textbf{Visualizing predictions.}
We further visualize the proclivities for \textbf{PRO} and \textbf{EXP} along with the true functions for both $w \in \{ w_{\rmss{exp}}, w_{\rmss{sig}} \}$ in Fig.~\ref{f:memfuncs}a.
While \textbf{True} and \textbf{EXP} do not learn speaking proclivity, we can illustrate the influence of the time-dependent term in~\eqref{eq:score} for each method by rescaling its proclivity by $\hat{\mbE}[d_i] / \hat{\mbE}[\pi_i]$, that is, the ratio of the empirical expected values of the memory and inherent scores, which are computed over a grid of equidistant traits $\bbx \in [0.1,1]^{50}$.
For $w_{\rmss{exp}}$ we find that the rescaled proclivities of \textbf{PRO} and \textbf{EXP} are very similar, as suggested by the similarity of their predictive losses in~\ref{f:boxplots}a.
However, for the more challenging $w_{\rmss{sig}}$, we find that \textbf{PRO} obtains a similarly decaying function to the true proclivity, albeit with a smaller effect for initial turns.
Indeed, by our definition of $w_{\rmss{sig}}$, speakers may not experience a noticeable decay in speaking likelihood until approximately $10$ turns elapse.
Conversely, \textbf{EXP} learns to reduce dependence on time by returning a smaller ratio $\mbE[\hat{d}_i]/\mbE[\hat{\pi}_i]$, hence its lower rescaled proclivity in~Fig~\ref{f:memfuncs} for all $\delta \geq 2$.
While this mitigates the effect of its incorrectly assumed proclivity $\hat{w}^{\rmss{EXP}}(\delta) = e^{-\delta / 2} \neq w_{\rmss{sig}}$, it also renders a worse predictive ability for \textbf{EXP}, as observed in Fig.~\ref{f:boxplots}b.

\subsection{Real-world Data}
\label{Ss:real}

Finally, we repeat the above experiments for real-world conversational data, described in greater detail in~\cite{obryan2025MLSPEAK}.
In brief, we observe a set of $G = 20$ teams of students among whom we observe conversations over Zoom, in which they discussed course projects.
Furthermore, we also have measurements of each student's level of extraversion $\{\bbx^{(g)} \}_{g=1}^{20}$, a personality trait that has a well-studied tendency to correlate with interactive behavior~\cite{macht2014StructuralModelsExtraversion,leung2001InterpersonalCommunicationPersonality}. 
We create 20 trials by first randomly selecting $12$ training, $4$ validation, and $4$ testing groups for the first trial.
Then, each subsequent trial corresponds to rotating group indices by one, such that every group is present or absent in training, validation, or testing sets for at least one trial.
As we no longer have ground truth likelihoods $\ccalU^*$, we compare test losses \textbf{PRO}, \textbf{EXP}, \textbf{NM}, and \textbf{HM} using predicted scores $\hbpi^{*(g)}, \hbd^{*(g)}$ of each testing group along with the proclivity $\hat{w}$ for each method.

Boxplot comparisons of test losses $\ell$ and $\ell_{\rmss{turn}}$ are shown in Fig.~\ref{f:boxplots}c.
We find that \textbf{PRO} is competitive with \textbf{EXP} and \textbf{HM} for the unweighted loss $\ell$.
As mentioned previously, real-world conversations often contain large numbers of floor turns.
Thus, methods such as \textbf{EXP} and \textbf{HM} can obtain low values of $\ell$ since they assume an exponentially decaying proclivity $w_{\rmss{exp}}$, which tends to encourage predicting floor turns.
Aligning with our intuition, \textbf{NM}, which does not employ time-varying predictions, yields the highest $\ell$ for the real-world conversations.
However, when we observe $\ell_{\rmss{turn}}$, for which lower values indicate greater ability to predict more complex turn-taking patterns, all methods \textbf{EXP}, \textbf{NM}, and \textbf{HM} attain significantly worse values than \textbf{PRO}.
Thus, our approach \textbf{PRO}, which learns time-varying proclivity, indeed exhibits flexibility when adapting to real-world conversational data, as we are able to demonstrate superior predictive ability for non-trivial turn-taking patterns.

Moreover, we visualize proclivities for the real-world data in Fig.~\ref{f:memfuncs}b, with $\hat{w}^{\rmss{EXP}}$ and $\hat{w}^{\rmss{PRO}}$ scaled by inherent and memory scores as described previously.
As we do not have ground truth proclivities, we instead plot $w_{\rmss{exp}}$ rescaled to compare its shape to $\hat{w}^{\rmss{PRO}}$, as $w_{\rmss{exp}}$ is a common choice for proclivity in past works~\cite{stasser1991Speakingturnsfacetoface,obryan2025MLSPEAK,obryan2022ConversationalTurntakingStochastic}.
First, we note that \textbf{EXP} reduces the effect of $\hat{w}^{\rmss{EXP}}$, as Fig.~\ref{f:memfuncs}b shows lower values of its scaled proclivity.
This implies that temporal effects are more subdued, potentially indicating that \textbf{EXP} employs an incorrect proclivity, which we also observed for the synthetic case $w = w_{\rmss{sig}}$.
For our method \textbf{PRO}, our time-dependent likelihoods decay more slowly than $w_{\rmss{exp}}$, where $\hat{w}^{\rmss{PRO}}$ yields much higher values in Fig~\ref{f:memfuncs}b than both $w_{\rmss{exp}}$ and $\hat{w}^{\rmss{EXP}}$, especially when $\delta$ is approximately between $5$ and $10$.
Thus, we find that these real-world conversations did not exhibit the typically assumed proclivity $w_{\rmss{exp}}$.
Overall, we find that our \textbf{PRO} model in~\eqref{eq:model} is able to identify time-dependent speaking behavior for real group interactions shown in Fig.~\ref{f:memfuncs}b, and we demonstrate its superior predictive ability for realistic, non-trivial speaking patterns in Fig.~\ref{f:boxplots}c in comparison with other methods.

\section{Conclusion}
\label{S:concl}

In this work, we presented a flexible computational model that predicts the likelihood that each group member will speak at a given speaking turn in a conversation based on their personality traits and conversation history.
Moreover, we introduced a novel approach to learn how speaking proclivity depends on how recently individuals have spoken, which allows our model to adapt to group settings with unique conversation dynamics.
In addition, we applied our model to real-world conversation data and compared the learned time-varying proclivities with previous assumptions of how speaking behavior evolves.
We found that the real-world group conversations that we observed exhibited different behavior than is commonly assumed, demonstrating the need to consider variation in group dynamics and also develop flexible tools to model them.
In future work, we will consider proclivity models more amenable to time-series predictions.

\bibliographystyle{IEEEbib}
\bibliography{citations}

\end{document}